\documentclass{article}

\PassOptionsToPackage{numbers}{natbib}
 \usepackage[preprint]{neurips_2026}

\usepackage[utf8]{inputenc} 
\usepackage[T1]{fontenc}    
\usepackage{hyperref}       
\usepackage{url}            
\usepackage{booktabs}       
\usepackage{amsfonts}       
\usepackage{nicefrac}       
\usepackage{microtype}      
\usepackage{xcolor}         
\usepackage{amsmath}
\usepackage{todonotes}
\usepackage{placeins}
\usepackage{subcaption}
\usepackage{graphicx}
\usepackage{adjustbox}
\usepackage{multirow}
\usepackage[normalem]{ulem}

\usepackage[dvipsnames,table]{xcolor}
\usepackage{etoolbox}
\robustify\cellcolor
\newcommand{\bgl}{\cellcolor[HTML]{DDDDDD}}
\newcommand{\ci}[1]{\ensuremath{\,{\scriptstyle \pm\,#1}}}

\newcommand{\second}[2]{%
  \underline{#1}\ci{#2}%
}

\title{Attribution via Distributional Paths for Information Revelation}

\author{
  Kieran A. Murphy \quad Shameen Shrestha \\[0.4em]
  New Jersey Institute of Technology \\[0.4em]
  \texttt{\{kieran.murphy, ss5375\}@njit.edu}
}

\begin{document}

\maketitle

\begin{abstract}

Feature attribution methods explain predictions by assigning importance scores to input features.
Path-based methods such as Integrated Gradients are especially appealing because they satisfy \textit{completeness}: attributions sum to the change in model output between a reference state and the input.
Yet most path methods define this trajectory in input space, explaining a model through pointwise perturbed inputs along a chosen path.
An input-space path integrates the model's raw response at each point it passes through, with no control over the resolution at which a feature is queried; the early, baseline-adjacent part of the trajectory contributes to the explanation on equal footing with the input itself.
Here, we lift path attribution from input space to a space of structured probe distributions around the example of interest, and call our method Reveal-IG.
Rather than traversing raw input values, Reveal-IG progressively reveals information about the input and attributes changes in the model's expected output along this distributional path.
The result is a path-attribution framework that retains completeness with respect to the expected model response, and naturally accommodates multiscale image probes and feature-wise uncertainty in tabular data.
Synthetic diagnostics show that Reveal-IG avoids path artifacts that affect input-space methods, and across ImageNet classification and tabular regression it produces stable, signed attributions---leading on metrics that use attribution sign while remaining competitive on the rest.
  
\end{abstract}

\section{Introduction}

\begin{figure}
    \centering
    \includegraphics[width=\linewidth]{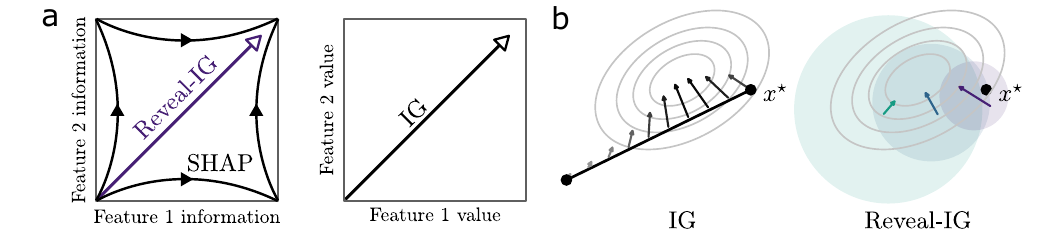}
    \caption{\textbf{(a)} SHAP reveals feature information in discrete steps, averaging contributions along all such paths. 
    We introduce Reveal-IG, a method that gradually reveals feature information, traversing a continuous path analogous to the path of Integrated Gradients (IG) in the space of feature values. 
    \textbf{(b)} IG evaluates along a single path from a baseline to the point being explained, $x^\star$.
    Reveal-IG integrates over a sequence of distributions that progressively concentrate around the input, capturing contributions across multiple locations and scales.}
    \label{fig:fig1}
\end{figure}

Explainable AI is an important component of modern machine learning systems, and feature attribution is one of its most widely used forms, assigning importance scores to input features in order to explain a model's prediction~\cite{samek2021xai, molnar2022interpretableML}. 
Two central approaches are Shapley Additive exPlanations (SHAP)~\cite{lundberg2017shap} and Integrated Gradients (IG)~\cite{sundararajan2017ig}. 
Both satisfy completeness: attributions sum to the change in model output between a reference state and the input being explained. 
Both can be understood as path-based methods (Fig.~\ref{fig:fig1}).
SHAP attributes to each feature its average contribution along discrete paths that iteratively reveal feature values, moving from a global reference distribution to the observed input. 
In contrast, IG decomposes a continuous path integral, typically along a trajectory in input space.

Despite this shared foundation, SHAP and IG expose distinct limitations. 
SHAP requires averaging over combinatorially many feature orderings, and its discrete reveal operation treats each feature as either unknown or fully observed. 
This discontinuous reveal operation is a severe intervention: a feature jumps from unobserved to fully observed, rather than allowing evidence to enter the explanation gradually.
IG provides a continuous decomposition, but explains the model through a trajectory of individual input points, which may pass through regions of input space that are not semantically meaningful and reflect only pointwise evaluations of the model~\cite{sturmfels2020distill}. 
Several variants modify this trajectory, for example by replacing the straight-line path with a scale-space path~\cite{Xu2020blurig} or with an adaptive path chosen to reduce noisy attributions~\cite{kapishnikov2021guidedIG}. 
These methods improve the choice of path, but the model is still evaluated pointwise along a trajectory of inputs.

Crucially, completeness is a property of the path integral, not of the particular space in which the path is drawn. 
It only requires a trajectory connecting a reference state to the explained prediction, along which changes in the model response can be aggregated and then decomposed into feature contributions. 
This opens the possibility of taking the path integral in a lifted space of structured probes rather than in the original input space. 
In this work, we instantiate this idea by defining continuous paths through distributional probes of the model: transformations that aggregate the model's behavior over structured perturbations of the input. 
In practice, these probes are implemented as expectations over parameterized distributions, enabling efficient evaluation via Monte Carlo sampling and reparameterization. 
We call the resulting method Reveal-IG, reflecting the gradual revelation of information about the datapoint being explained within an IG-style attribution objective.
The result is a continuous analogue of SHAP-style feature revelation, replacing discrete feature coalitions with a differentiable reveal path from broad probes to probes concentrated around the explained input.

This lifted view gives a common language for attribution across domains. 
In each case, attribution is defined by a path through structured probe distributions---from a broad, high-uncertainty reference to a probe concentrated on the observed example---that specifies not only which feature values are shown to the model but the resolution at which they are queried. 
The probe family is the only domain-specific choice: it adapts to the structure of images or tabular data while the path-integral construction stays fixed. 
We instantiate Reveal-IG for image classification and tabular regression, where it yields interpretable attribution trajectories and competitive-to-strong performance against gradient-, perturbation-, and Shapley-style baselines.

\section{Method}
Consider a model $f$ and an input $x^\star$ whose prediction we wish to explain. 
We index the features of $x^\star$ by $i$, and seek feature-wise attributions that decompose the change in model response between a reference state and the explained input.

Path-based attribution methods formalize such decompositions by applying the
fundamental theorem of calculus along a path. Let $\theta$ denote the coordinates
being varied along the path, and let $G(\theta)$
be the scalar model quantity whose change we wish to decompose. 
For a path
$\mathcal{S}$ from $\theta^{\mathrm{start}}$ to $\theta^{\mathrm{end}}$,
\begin{equation}
\label{eqn:path}
    G(\theta^{\mathrm{end}}) - G(\theta^{\mathrm{start}})
    =
    \int_{\mathcal{S}} \nabla_\theta G(\theta) \cdot d\theta
    =
    \sum_i \int_{\mathcal{S}}
    \frac{\partial G}{\partial \theta_i} \, d\theta_i = \sum_i a_i.
\end{equation}
The terms $a_i$ define the path-dependent, per-feature attributions, where $\theta_i$ denotes the block of path coordinates associated with feature $i$ and the sum runs over features.

In standard input-space formulations, the path coordinates are the input values themselves: $\theta=x$ and $G(\theta)=f(x)$, so each feature's block $\theta_i$ is the single scalar $x_i$, and the attribution decomposes the change in model output as the input moves from a baseline $x^{\mathrm{start}}$ to the example $x^{\mathrm{end}}=x^\star$.
In our formulation, by contrast, $\theta$ parameterizes a family of probe
distributions $q_\theta(z \mid x^\star)$, and
\begin{equation}
        G(\theta;x^\star) = \mathbb{E}_{z \sim q_\theta(\cdot \mid x^\star)}[f(z)] .
\end{equation}
The path-integral decomposition is applied to a path through the space of distributions over inputs, rather than through input space directly.
Attribution is obtained by integrating gradients of this expected response along the path of probe distributions:
\begin{equation}
\label{eqn:klig}
    \Delta G
    =
    \int_{\mathcal{S}} 
    \nabla_\theta \mathbb{E}_{z \sim q_\theta(z \mid x^\star)}[f(z)]
    \cdot d\theta .
\end{equation}

For practical implementation, we assume a factorized form for the probe distribution,
\begin{equation}
q_\theta(z \mid x^\star) = \prod_i q_{\theta_i}(z_i \mid x_i^\star).
\end{equation}
When a feature is represented by multiple probe parameters, its attribution is the sum of the corresponding line-integral terms.
For Gaussian probes, each feature is represented by a parameter block
$\theta_i=(\mu_i,\ell_i)$, where $\ell_i=\log\sigma_i^2$.
The attribution to feature $i$ is the line-integral contribution of this block:
\begin{equation}
a_i
=
\int_0^1
\left[
\frac{\partial G}{\partial \mu_i}\frac{d\mu_i}{dt}
+
\frac{\partial G}{\partial \ell_i}\frac{d\ell_i}{dt}
\right]dt .
\end{equation}
Thus the reported feature attribution is the sum of the mean-path and variance-path contributions; this sum is required by completeness and is not an additional weighting heuristic.

Under this parameterization, the key design choice is the trajectory of feature-wise probe distributions. We define a path that begins at a broad reference probe and ends at a probe concentrated around the example $x^\star$. 
The framework does not require it, but in both of our instantiations the reference is tied to the data distribution: it is the factorized marginal of $p_{\mathrm{data}}$, realized as a moment-matched Gaussian for images and as the empirical marginal for tabular features. 
This makes the start of the path a maximum-uncertainty state under the factorized approximation, and grounds the reference in the population.
The endpoint is a probe concentrated on $x^\star$---a product of finite-variance distributions
---corresponding to near-full specification of the explained input.

\paragraph{Image data.}
For image data, we define a local probe distribution around the explained image $x^\star$ by modeling each color channel of each pixel as an independent Gaussian variable. 
Let $z$ denote an image sampled from the probe. 
We parameterize
\begin{equation}
    q_\theta(z \mid x^\star)
    =
    \prod_i \mathcal{N}(z_i; \mu_i, \sigma_i^2),
\end{equation}
where $i$ indexes pixel-channel components. 
The path begins at a reference probe with $\mu_i=0$ and $\sigma_i=1$, corresponding to the normalized image-space prior, and ends at a probe centered on the observed image, with $\mu_i=x_i^\star$ and $\sigma_i=\sigma_{\mathrm{final}}$.

We consider both fixed and adaptive choices of $\sigma_{\mathrm{final}}$. 
In the fixed setting, the endpoint retains a constant amount of residual uncertainty around each pixel value. 
In the adaptive setting, $\sigma_{\mathrm{final}}$ is chosen separately for each image, based on the scale of additive Gaussian noise under which the expected target logit remains above a specified fraction of its clean value.
This allows the endpoint probe to reflect the resolution at which the model remains sensitive to perturbations of the input.

\paragraph{Tabular data.}
For tabular data, we do not assume a dense input space from which meaningful perturbations can be sampled freely.
Instead, we define probe distributions over empirical feature values observed in the training data.
For each feature, we construct a distribution that transitions from the empirical marginal distribution to a concentrated distribution around the observed value $x_i^\star$.

We parameterize each feature-wise probe using a temperature-controlled similarity kernel,
\begin{equation}
\label{eqn:tabular_dists}
    q_{\tau_i}(z_i \mid x_i^\star)
    \propto 
    p_{\mathrm{data}}(z_i)
    \exp\!\left(s_i(z_i, x_i^\star) / \tau_i\right),
\end{equation}
where $z_i$ denotes a candidate value for feature $i$, $p_{\mathrm{data}}(z_i)$ is the empirical marginal distribution, and $s_i(\cdot,\cdot)$ is a feature-specific similarity function. 
In our experiments, all features are treated as numeric, and we use negative squared distance as the similarity function.
Other empirical kernels, such as equality kernels for categorical variables, can be used but are not explored here.

The path is defined by decreasing the temperatures $\tau_i$, which progressively concentrates each feature-wise probe around the observed value $x_i^\star$. 
At high temperature, each probe approaches the empirical marginal distribution; at low temperature, it approaches a near-deterministic distribution at the observed feature value. 
Expectations under the resulting product probe are estimated using empirical feature values from the training data under a factorized, interventional approach.
That is, feature values are recombined across datapoints, as in interventional SHAP~\cite{lundberg2017shap,covert2020sage}, with feature-wise importance weights determined by Eqn.~\ref{eqn:tabular_dists}.
We further allow feature-wise temperature schedules parameterized by fractional entropy reduction, so that features with different marginal entropies reveal information at matched rates and reach near-deterministic endpoint distributions simultaneously.
This prevents low-entropy features from being effectively revealed earlier than high-entropy features simply because their marginal distributions are already concentrated.

\paragraph{Practical evaluation of gradients.}
The gradient in Eqn.~\ref{eqn:klig} is taken with respect to the probe parameters $\theta$, not directly with respect to the input values. 
We evaluate these gradients using standard estimators for differentiating expectations.

When the probe distribution admits a differentiable reparameterization, we write samples as $z=g_\theta(\epsilon;x^\star)$ for fixed noise $\epsilon$, and estimate gradients by differentiating $f(g_\theta(\epsilon;x^\star))$ with respect to $\theta$ using automatic differentiation~\cite{vae}. 
For Gaussian image probes, this corresponds to sampling $z_i=\mu_i+\sigma_i\epsilon_i$ with $\epsilon_i\sim\mathcal{N}(0,1)$.
For tabular data, where probes are defined over empirical feature values, we estimate expectations using training-set feature values with weights induced by the feature-wise probe distributions.
The empirical feature values are fixed and only the weights depend on the probe parameters; differentiating the weighted expectation therefore estimates the gradient of the expected model response with respect to the probe parameters.
This yields a black-box estimator requiring only model evaluations, making the method applicable to tree ensembles and other non-differentiable predictors.

\begin{figure}
    \centering
    \includegraphics[width=\linewidth]{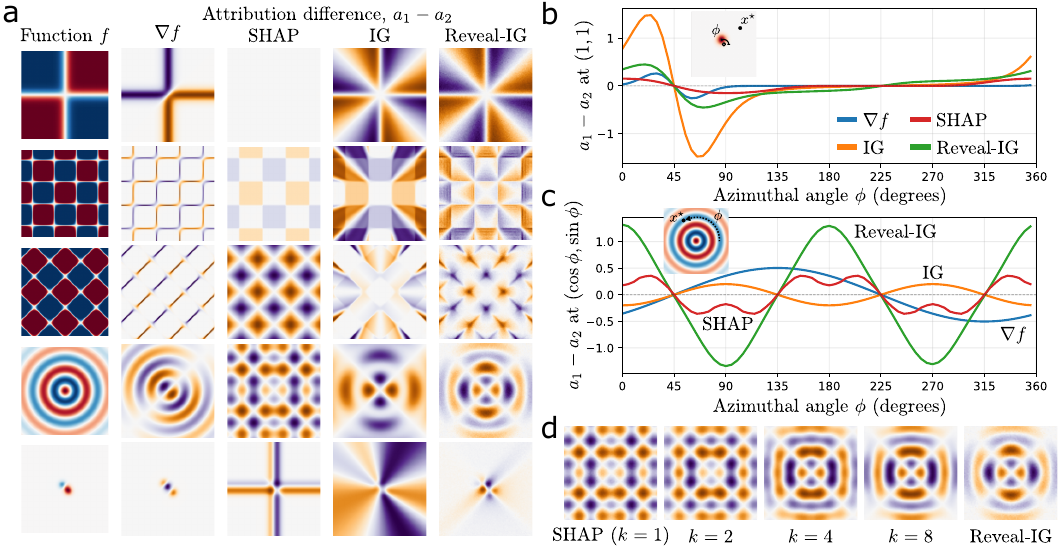}
    \caption{Attribution fields in two dimensions.
    \textbf{(a)} For a selection of functions (left column), we evaluate attributions according to gradients, SHAP, IG, and Reveal-IG.
    The difference of attribution components, $a_1-a_2$ is displayed as a heatmap.
    \textbf{(b)} By systematically varying the function, we obtain response curves for the attribution methods.
    A bump in the function's output is rotated around the origin and the attribution at $(1,1)$ is recorded.
    \textbf{(c)} For the radial rings in panel \textbf{a}, we extract the attribution difference around a circle of radius one.
    \textbf{(d)} By averaging over paths with $k$ steps of feature information reveal, the radial rings' attribution interpolates from SHAP toward Reveal-IG.
    }
    \label{fig:synthetic_comparison}
\end{figure}

\section{Experiments}
\subsection{Synthetic diagnostics: attribution fields in two dimensions}
\label{sec:synthetic}

We begin with scalar functions on a two-dimensional input space (Fig.~\ref{fig:synthetic_comparison}a), where both the input-output landscape and the resulting attribution field can be visualized directly.
These examples are not benchmarks; they are controlled diagnostics for how attribution methods respond to structure away from the point being explained.
For complete two-feature methods, the attribution vector $(a_1,a_2)$ satisfies
$a_1+a_2=f(x^{\mathrm{end}})-f(x^{\mathrm{start}})$.
We visualize the remaining degree of freedom, $a_1-a_2$, which shows how each method allocates the same total change across the two coordinates.
For gradients, which have two free components, we likewise plot their difference.

The first rows of Fig.~\ref{fig:synthetic_comparison}a show some agreement in simple cases that diminishes as the landscape becomes more structured.
For the XOR-like function, IG and Reveal-IG produce closely related attribution fields.
For the checkerboard functions, both methods highlight sharp changes near corners and edges, though with different emphasis induced by their respective paths.
SHAP produces more axis-aligned structure, particularly visible in the last two rows, reflecting that its explanations are built from coordinate-wise feature reveal operations rather than continuous motion through the input or distribution space.

The final row highlights a sharper failure mode of input-space paths.
IG produces ``shadows'': attribution at a point can be strongly influenced by features encountered along the straight-line path from the baseline, even when those features are far from the point being explained.
Here, localized structure in the function's response near the origin casts long imprints through the attribution field.
Reveal-IG shows a much less pronounced effect because the early, large-length scale probes average model responses over a broad region of the landscape and dilute the influence of distant structure before the distributions concentrate near $x^\star$.

We isolate this shadow effect using a controlled function sweep in Fig.~\ref{fig:synthetic_comparison}b.
A localized bump is moved around a circle centered at the origin, while the attribution at a fixed point $x^\star$ is measured.
Although the bump remains far from $x^\star$, IG changes substantially when the bump crosses the straight-line path from the baseline to $x^\star$.
Reveal-IG shows a weaker, though nonzero version of the same effect.

Finally, Fig.~\ref{fig:synthetic_comparison}c examines angular structure in attributions when the underlying function is radially symmetric.
Gradients show first-harmonic dependence while the complete methods are dominated by the second harmonic.
SHAP contains additional higher-order angular components.
We test whether the severity of SHAP's revelation operation---a single all-or-nothing step---is responsible for the axis-aligned structure by subdividing each feature's reveal into $k$ equal-uncertainty steps and averaging over the possible orderings.
$k=1$ is equivalent to SHAP, and $k \to \infty$ yields a path-ensembling attribution that approaches the continuous reveal path underlying Reveal-IG. 
As $k$ increases, higher-order angular components diminish toward the nonzero value retained by Reveal-IG itself (Fig.~\ref{fig:synthetic_comparison}d). 
The severity of the reveal step thus has a measurable effect on attributions, independent of whether that effect is desirable; across functions, intermediate $k$ produces attribution maps that interpolate between the SHAP and Reveal-IG extremes (Appx.~\ref{appx:extended_synth}).

Together, these diagnostics show what is hard to see in high dimensions: IG can inherit artifacts from a single input-space path, SHAP can introduce axis-aligned reveal structure, and Reveal-IG presents a middle-ground that partially mitigates both effects.

\begin{figure}
    \centering
    \includegraphics[width=\linewidth]{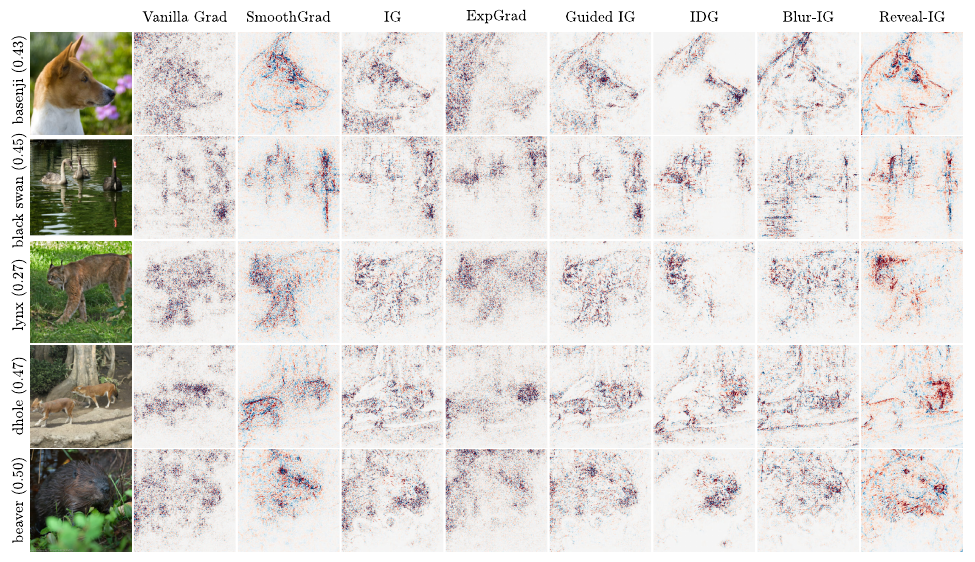}
    \caption{For a random selection of ImageNet validation images, we show signed saliency maps for the methods evaluated in Tab.~\ref{tab:image}. The left column shows the input image together with the ResNet-50 top-1 prediction and confidence. Red is positive attribution and blue is negative.
    }
    \label{fig:comparison_resnet}
\end{figure}

\subsection{Image classification}
\label{sec:image_results}

\paragraph{Data and model.}
We evaluate image attributions on 1,000 images sampled from the ImageNet-1K validation set~\citep{deng2009imagenet,russakovsky2015imagenet}, using a pre-trained ResNet-50~\cite{he2016deep} with \texttt{IMAGENET1K\_V2} weights.
We repeat the quantitative image metrics on ViT-B/16~\cite{dosovitskiy2021vit} (\texttt{IMAGENET1K\_SWAG\_LINEAR\_V1} weights) to test whether the trends persist beyond convolutional architectures.
Inputs are resized to 256 px, center-cropped to $224\times224$, and normalized with ImageNet statistics.
Unless otherwise stated, the attribution target for image experiments is the ground-truth class if its predicted probability exceeds 0.05, and the model’s top-1 predicted class otherwise.

\begin{table*}[t]
\centering
\caption{Image attribution across architectures (95\% CIs).
Insertion/deletion and Sensitivity-$n$ are evaluated on $n=1{,}000$ ImageNet validation images, sampled as one image per class.
Insertion/deletion uses a Gaussian-blur baseline following the RISE protocol;
\emph{Ins-Del$|$abs}: pixels ranked by $|\mathrm{attr}|$;
\emph{Ins-Del$|$relu}: pixels ranked by $\mathrm{relu}(\mathrm{attr})$ (positive evidence only).
RMA is evaluated on $n=879$ ImageNet-S validation images with ground-truth segmentation masks for ResNet-50 ($n=914$ for ViT due to higher accuracy).
Shading indicates best, \underline{underline} second best.}
\label{tab:image}
\renewcommand{\arraystretch}{1.2}
\setlength{\tabcolsep}{5pt}
\small
\begin{tabular}{lrrrr}
\toprule
\textbf{Method} & \textbf{Ins-Del$|$abs $\uparrow$} & \textbf{Ins-Del$|$relu $\uparrow$} & \textbf{RMA $\uparrow$} & \textbf{Sens.-$n$ $\uparrow$} \\
\midrule
\multicolumn{5}{l}{\textit{ResNet-50}} \\
\midrule
Vanilla Grad & 0.0206\ci{0.0056} & 0.0042\ci{0.0037}  & 0.4884\ci{0.0184}  & -0.0036\ci{0.0039} \\
SmoothGrad & 0.1652\ci{0.0076} & 0.0195\ci{0.0048} & 0.5672\ci{0.0163} & -0.0014\ci{0.0040} \\
IG & 0.0798\ci{0.0063} & 0.0760\ci{0.0048}  & 0.5177\ci{0.0179} & 0.0203\ci{0.0039} \\
ExpGrad & 0.0849\ci{0.0063} & 0.0858\ci{0.0047} & 0.5047\ci{0.0170} & 0.0137\ci{0.0039} \\
Guided IG & 0.1329\ci{0.0069} & \second{0.1043}{0.0053}  & 0.5787\ci{0.0164} & \second{0.0358}{0.0038} \\
IDG & 0.1772\ci{0.0078} & 0.1003\ci{0.0055}  & \bgl{0.6139}\ci{0.0161} & 0.0210\ci{0.0041} \\
Blur-IG & \bgl{0.1987}\ci{0.0078} & 0.0847\ci{0.0044} & 0.5663\ci{0.0168} & 0.0032\ci{0.0039} \\
Reveal-IG & \second{0.1960}{0.0081} & \bgl{0.2343}\ci{0.0091} & \second{0.5956}{0.0153} & \bgl{0.0802}\ci{0.0043} \\
\midrule
\multicolumn{5}{l}{\textit{ViT-B/16}} \\
\midrule
Vanilla Grad & 0.0725\ci{0.0119} & 0.0213\ci{0.0073} & 0.4205\ci{0.0173}  & -0.0025\ci{0.0040} \\
SmoothGrad & 0.3263\ci{0.0140} & 0.0526\ci{0.0092}  & 0.5283\ci{0.0157} & -0.0059\ci{0.0040} \\
IG & 0.1641\ci{0.0124} & 0.2090\ci{0.0105}  & 0.4709\ci{0.0163} & 0.0226\ci{0.0040} \\
ExpGrad & 0.1493\ci{0.0113} & 0.1980\ci{0.0092}  & 0.4442\ci{0.0164} & 0.0138\ci{0.0039} \\
Guided IG & 0.2249\ci{0.0127} & 0.2426\ci{0.0110}  & 0.5139\ci{0.0154}  & \second{0.0335}{0.0039} \\
IDG & 0.2800\ci{0.0133} & 0.2604\ci{0.0117}  & \second{0.5359}{0.0157} & 0.0246\ci{0.0041} \\
Blur-IG & \second{0.3658}{0.0130} & \second{0.2689}{0.0103} & 0.5238\ci{0.0154} & 0.0113\ci{0.0040} \\
Reveal-IG & \bgl{0.3774}\ci{0.0140} & \bgl{0.4951}\ci{0.0146} & \bgl{0.5545}\ci{0.0147} & \bgl{0.0808}\ci{0.0043} \\
\bottomrule
\end{tabular}
\end{table*}

\paragraph{Attribution methods.}
We compare Reveal-IG against a set of gradient- and path-based baselines: Guided IG~\citep{kapishnikov2021guidedIG}, IG with a zero baseline~\citep{sundararajan2017ig}, Integrated Decision Gradients (IDG)~\citep{walker2024idg}, Blur-IG~\citep{Xu2020blurig}, Expected Gradients~\citep{erion2021expgrad}, SmoothGrad~\citep{smilkov2017smoothgrad}, and Vanilla Gradient~\citep{simonyan2013deep}.
Reveal-IG is evaluated with a fixed terminal scale $\sigma=0.25$; we explore other endpoints and an adaptive-$\sigma$ variant that chooses the largest terminal scale preserving at least 95\% of the clean target logit under Gaussian perturbations in Appx.~\ref{appx:ablations}.
Both variants use 50 integration steps with 10 Monte Carlo samples per step.
All methods return signed channel-level attributions, which are collapsed by summing over channels.
Full hyperparameters are given in Appendix~\ref{app:impl}.

\paragraph{Evaluation protocol.}
We use three complementary families of metrics.
First, insertion and deletion measure whether highly ranked pixels have large effects on the target logit when pixels are progressively restored from, or removed to, the same blurred baseline~\cite{petsiuk2018rise} (blur $\sigma=10$).
We report both magnitude-based ranking, $|\mathrm{attr}|$, which treats positive and negative evidence as saliency, and positive-evidence ranking, $\text{ReLU}(\mathrm{attr},0)$, which asks specifically whether the method identifies features supporting the target class.
Second, Sensitivity-$n$ evaluates whether the signed attribution assigned to a random subset of pixels predicts the signed change in target logit when that subset is removed~\cite{ancona2018sensitivityn}.
Third, object-concentration metrics measure how much attribution mass falls within externally defined object regions.
We report Relevance Mass Accuracy (RMA)~\cite{arras2022clevr}, the fraction of positive attribution mass inside the object mask, using ImageNet-S ground-truth segmentation masks~\cite{gao2022imagenets}.
RMA is evaluated only on examples for which the foreground mask covers more than 2\% of the image. 
This separation is important: insertion/deletion tests faithfulness under a specific perturbation procedure, Sensitivity-$n$ tests signed additive consistency, and RMA tests spatial alignment with semantic foreground.

\paragraph{Image attribution results.}
The quantitative results are summarized in Tab.~\ref{tab:image}, with representative saliency maps shown in Fig.~\ref{fig:comparison_resnet}.
Across both architectures, Reveal-IG is best or second-best on every reported metric.
Its largest gains appear on evaluations that use attribution sign.
Positive-evidence insertion/deletion uses sign to select target-supporting pixels, while Sensitivity-$n$ evaluates the attribution map as a signed decomposition by comparing signed attribution sums to signed target-logit changes under feature perturbations.
Reveal-IG leads both metrics by a wide margin, suggesting that its attributions better distinguish target support from target suppression.

This interpretation is consistent with Fig.~\ref{fig:comparison_resnet}, where Reveal-IG shows clearer separation between positive and negative evidence than the baselines.
When signs are discarded, the gap narrows: magnitude-ranked insertion/deletion treats positive and negative evidence together as saliency. 
By contrast, the largest Reveal-IG gains appear when the evaluation uses attribution sign, either by ranking positive evidence or by comparing signed attribution sums to signed output changes.
Under these unsigned evaluations, Reveal-IG remains near-best, but Blur-IG and IDG are stronger competitors; qualitatively, magnitude visualizations also make the methods appear more similar (Appx.~\ref{appx:extra_vis}).
Thus Reveal-IG's main advantage is not simply producing more concentrated saliency maps, but assigning signs that better correspond to target support and suppression.

The closest competitors emphasize different strengths.
Blur-IG is strongest for magnitude-ranked insertion/deletion, showing that a coarse-to-fine image path can be highly effective for unsigned saliency.
IDG is strongest for ResNet-50 localization, indicating that its path can produce spatially concentrated maps.
Both Blur-IG and Reveal-IG replace direct point-to-point interpolation with paths organized around progressive information revelation, but they smooth different objects: Blur-IG smooths the input space by progressively restoring spatial detail, whereas Reveal-IG smooths the output response by integrating predictions over distributions that progressively concentrate around the observed image.
\begin{figure}
   \centering
   \includegraphics[width=\linewidth]{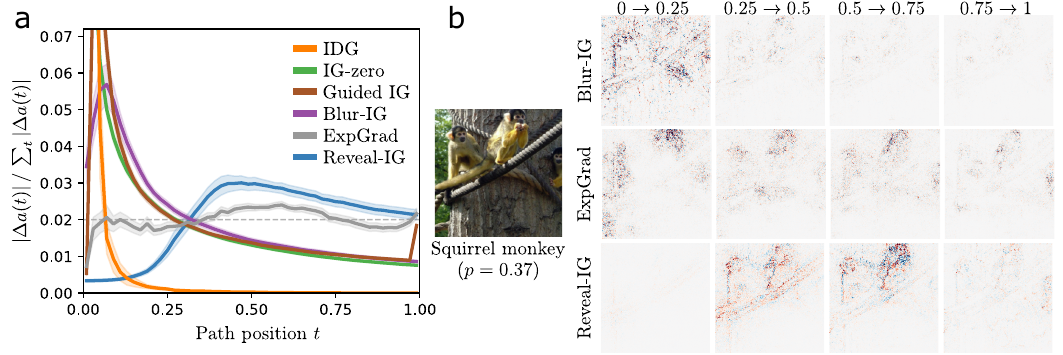}
   \caption{Attribution over the path.
   \textbf{(a)} We show the fractional rate of change of the attributions, $|\Delta a(t)| / \sum_t |\Delta a(t)|$, averaged over 100 images.
   \textbf{(b)} Contribution to final attribution maps across segments of the trajectory for a randomly selected image.}
   \label{fig:path}
\end{figure}

\paragraph{Attribution along the path.}
To examine how different path methods distribute attribution across the trajectory, Fig.~\ref{fig:path} plots the fraction of total absolute attribution accumulation as a function of path progress, averaged over 100 validation images.
The normalization forces each curve to sum to one, so the plot compares where along the path attribution is assigned rather than the total magnitude or completeness of the attribution.
The methods form four natural groups.
IDG is the most concentrated, with attribution spiking at the beginning of the path; this reflects its decision-region weighting, which emphasizes regions where the target logit changes rapidly rather than preserving the standard unweighted IG integral.
IG, Guided IG, and Blur-IG also assign the majority of their attribution early in the trajectory, indicating that much of their attribution is accumulated while the path is still far from the input.
Expected Gradients allocates attribution nearly uniformly along the trajectory, while the Reveal-IG variants assign most attribution late in the path.
This late accumulation is more consistent with the reveal-based interpretation: attribution is assigned as the probe becomes informative about the explained input, rather than being dominated by regions near the baseline.
It also echoes the synthetic diagnostics, where early path dependence produced shadow artifacts (Fig.~\ref{fig:synthetic_comparison}); delaying attribution until the probe concentrates near the explained image reduces the opportunity for distant or baseline-adjacent structure to dominate the explanation.

\subsection{Tabular regression}
\label{sec:tabular}

\paragraph{Data and model.}
We evaluate tabular attributions on three regression datasets: Bike Sharing Demand, California Housing, and Wine Quality.
Bike Sharing Demand is the UCI Bike Sharing dataset, accessed through OpenML (data ID 42712)~\cite{fanaee2014event}.
California Housing is the StatLib California housing dataset, accessed through \texttt{sklearn}~\cite{pace1997sparse}.
Wine Quality is the combined red and white UCI Wine Quality dataset, accessed through OpenML (data ID 287)~\cite{cortez2009modeling}.
For each dataset, we train a two-hidden-layer MLP regressor with 64 hidden units per layer and ReLU activations, and compute attributions with respect to the scalar model prediction.
All features are standardized using training-set statistics, and regression targets are standardized before training.
Unless otherwise stated, tabular results are evaluated on 1,000 held-out test examples.

\paragraph{Attribution methods.}
We compare Reveal-IG against gradient-, path-, perturbation-, and local-surrogate baselines: Vanilla Gradient, SmoothGrad, IG, Expected Gradients, KernelSHAP~\citep{lundberg2017shap}, LIME~\citep{ribeiro2016should}, and feature ablation.
For Reveal-IG, feature-wise probe distributions are defined over empirical marginal feature values and annealed toward the observed feature values along a normalized entropy path.
The path begins near the empirical marginal distribution, at $0.99$ of each feature's marginal entropy, and ends at a concentrated probe with $0.05$ of that entropy.
For IG, the reference point is the training-set mean after standardization, i.e., zero.
Feature ablation replaces each feature with samples from its empirical marginal distribution and averages the resulting prediction change. 
Full hyperparameters are given in Appendix~\ref{app:impl}.

\paragraph{Evaluation protocol.}
Tabular data is a challenging setting for input-space path methods because the geometry of the input space is not obvious.
Features can have different scales and heterogeneous semantics, so linear interpolation between a baseline and an example can be hard to interpret and may pass through implausible or out-of-distribution inputs.
This makes tabular data a useful test of the information-revelation view: rather than walking through a single standardized feature space, Reveal-IG reveals feature information through empirical marginal probes.

We evaluate using sufficiency, comprehensiveness, sensitivity-max, and a signed directional insertion diagnostic.
Sufficiency and comprehensiveness measure whether the top-ranked features are respectively adequate to preserve the prediction and necessary for the original prediction, following \citet{deyoung2020eraser}.
We rank features by attribution magnitude and use the top 20\% of features. 
For sufficiency, we retain only these features and replace the rest using the training-set mean; lower values indicate that the selected features preserve the original prediction. 
For comprehensiveness, we replace the top-ranked features and measure the absolute change in prediction; higher values indicate that the selected features capture information the model relies on. 
Because the task is regression, both metrics are computed as absolute differences in the scalar model output.
Sensitivity-max measures local stability of the attribution map under small input perturbations~\citep{yeh2019fidelity}.
Unlike Sensitivity-$n$, which evaluates whether attribution sums over feature subsets match corresponding output changes, Sensitivity-max asks whether the explanation itself is stable in a local neighborhood of the input.
We estimate Sensitivity-max by sampling 50 random perturbation directions on the $\ell_2$ sphere of radius $r=0.1$ in standardized feature space, and report the largest normalized attribution change,
$
\max_m \frac{\|\Phi(x+\delta_m)-\Phi(x)\|_2}{r}.
$

The directional insertion diagnostic adapts insertion/deletion to signed regression attributions.
For tabular regression, the sign of the attribution directly predicts the direction in which the model output should move.
We therefore reveal features in descending attribution order and in ascending attribution order, compute the AUC of each prediction curve as a function of the fraction of features revealed, and report their difference, $\mathrm{AUC}_{\mathrm{desc}}-\mathrm{AUC}_{\mathrm{asc}}$.
A good signed attribution should produce a large separation: revealing positively attributed features first should raise the prediction, while revealing negatively attributed features first should lower it.

\begin{table*}[t]
\centering
\caption{Tabular attribution evaluation (95\% CIs). Shading indicates best, \underline{underline} second best.}
\label{tab:tabular_main}
\renewcommand{\arraystretch}{1.2}
\setlength{\tabcolsep}{5pt}
\small
\begin{tabular}{llrrrr}
\toprule
\textbf{Data} & \textbf{Method} & \textbf{Ins-Dir $\uparrow$} & \textbf{Suff.\ $\downarrow$} & \textbf{Comp.\ $\uparrow$} & \textbf{Sens.-max $\downarrow$} \\
\midrule
\multirow{8}{*}{Bike}
& Vanilla Grad & 0.098\ci{0.049} & 0.500\ci{0.034} & 0.675\ci{0.041} & 30.119\ci{2.638} \\
& SmoothGrad & 0.131\ci{0.049} & 0.496\ci{0.033} & 0.643\ci{0.040} & 13.063\ci{0.747} \\
& IG & 0.819\ci{0.044} & 0.514\ci{0.036} & 0.641\ci{0.040} & 6.441\ci{0.369} \\
& ExpGrad & 0.636\ci{0.050} & 0.486\ci{0.032} & 0.706\ci{0.041} & 38.358\ci{0.860} \\
& KernelSHAP & \bgl{1.010}\ci{0.044} & \second{0.378}{0.024} & \bgl{0.848}\ci{0.044} & 4.122\ci{0.093} \\
& LIME & 0.082\ci{0.039} & 0.570\ci{0.040} & 0.543\ci{0.030} & \bgl{2.012}\ci{0.021} \\
& Feat.\ Ablation & 0.881\ci{0.043} & 0.401\ci{0.025} & 0.801\ci{0.042} & 5.125\ci{0.185} \\
& Reveal-IG & \second{0.973}{0.046} & \bgl{0.367}\ci{0.024} & \second{0.817}{0.043} & \second{2.474}{0.137} \\
\midrule
\multirow{8}{*}{Calif.}
& Vanilla Grad & 0.017\ci{0.070} & 1.019\ci{0.063} & 1.063\ci{0.059} & 80.602\ci{3.451} \\
& SmoothGrad & -0.393\ci{0.043} & 0.905\ci{0.048} & \bgl{1.399}\ci{0.065} & 38.762\ci{1.183} \\
& IG & \second{2.229}{0.081} & 0.534\ci{0.039} & 1.011\ci{0.051} & 13.891\ci{0.635} \\
& ExpGrad & 2.125\ci{0.084} & 0.535\ci{0.033} & 1.053\ci{0.054} & 28.091\ci{0.663} \\
& KernelSHAP & \bgl{2.348}\ci{0.083} & \bgl{0.448}\ci{0.027} & 1.018\ci{0.050} & \second{8.582}{0.179} \\
& LIME & -0.677\ci{0.039} & 1.488\ci{0.087} & 1.075\ci{0.048} & 12.711\ci{0.130} \\
& Feat.\ Ablation & 2.037\ci{0.086} & 0.705\ci{0.042} & \second{1.397}{0.073} & 11.816\ci{0.359} \\
& Reveal-IG & 2.183\ci{0.088} & \second{0.525}{0.036} & 1.074\ci{0.054} & \bgl{6.790}\ci{0.347} \\
\midrule
\midrule
\multirow{8}{*}{Wine}
& Vanilla Grad & -0.587\ci{0.065} & 0.748\ci{0.045} & 0.761\ci{0.050} & 21.175\ci{0.687} \\
& SmoothGrad & -0.660\ci{0.069} & 0.699\ci{0.044} & 0.794\ci{0.050} & 7.555\ci{0.161} \\
& IG & 1.371\ci{0.065} & 0.704\ci{0.043} & 0.894\ci{0.057} & 5.493\ci{0.233} \\
& Expected Grad & 1.116\ci{0.064} & 0.726\ci{0.048} & 0.852\ci{0.058} & 18.808\ci{0.310} \\
& KernelSHAP & \bgl{1.505}\ci{0.068} & \bgl{0.638}\ci{0.042} & \second{1.008}{0.064} & 3.932\ci{0.059} \\
& LIME & -0.657\ci{0.056} & 0.773\ci{0.050} & 0.672\ci{0.046} & \second{2.968}{0.032} \\
& Feat.\ Ablation & 1.220\ci{0.062} & 0.681\ci{0.045} & \bgl{1.137}\ci{0.064} & 6.705\ci{0.125} \\
& Reveal-IG & \second{1.438}{0.065} & \second{0.665}{0.044} & 0.928\ci{0.061} & \bgl{2.188}\ci{0.102} \\
\bottomrule
\end{tabular}
\end{table*}

\paragraph{Tabular attribution results.}
The tabular results in Table~\ref{tab:tabular_main} are more mixed than the image results, but show that the distributional-path idea remains competitive outside image geometry.
Reveal-IG is consistently among the strongest methods on directional insertion and sufficiency, and achieves the best or second-best sensitivity-max score on all three datasets.
This stability advantage is especially notable relative to the other gradient and path-based methods: vanilla gradients and SmoothGrad often fail under directional insertion, while Expected Gradients can be highly sensitive despite reasonable sufficiency or comprehensiveness scores.

At the same time, Reveal-IG is not uniformly dominant on tabular perturbation metrics.
KernelSHAP and feature ablation remain highly competitive, as their explanations are themselves defined through feature replacement.
This makes sufficiency and comprehensiveness a favorable evaluation setting for perturbation-native methods.
Reveal-IG instead occupies a different tradeoff: it approaches the perturbation baselines on removal-style metrics while producing substantially more stable attributions.

Overall, the tabular experiments support the generality of Reveal-IG rather than the stronger dominance seen in images.
They also highlight why the distributional reveal path is better matched to tabular data than ordinary input-space IG: without an obvious spatial or geometric interpolation between a baseline and an example, revealing feature information through empirical marginal probes provides a more natural path than moving linearly through standardized feature space.
Additionally, whereas IG requires differentiability along a continuous input-space path, tabular Reveal-IG moves probability mass over empirical feature values. 
Thus the method preserves discrete feature semantics and can explain black-box tabular models such as XGBoost using the same formulation.

\section{Related work}

\paragraph{Path attribution and completeness.}
Integrated Gradients (IG) attributes a prediction by integrating gradients along a path from a reference input to the example being explained, yielding a complete decomposition of the change in model output~\cite{sundararajan2017ig}.
This path-integral view is closely related to Aumann--Shapley cost sharing~\cite{aumann1974values}, and has motivated a family of methods that preserve completeness while modifying the choice of reference point or path.
Several works study the sensitivity of IG to its baseline and propose alternative baseline choices~\cite{sturmfels2020distill}.
Others modify the path itself, for example by replacing the straight-line path with a scale-space path~\cite{Xu2020blurig,kim2026spectral}, an adaptive path chosen to avoid high-gradient regions~\cite{kapishnikov2021guidedIG}, or a path designed to emphasize decision-relevant regions~\cite{walker2024idg}.
These methods demonstrate that the path is not merely an implementation detail: it determines which regions of the model's input-output landscape are incorporated into the explanation.

\paragraph{Stochastic and distributional path variants.}
A related line of work incorporates stochasticity into attribution by averaging over baselines, paths, or perturbations.
Expected Gradients averages IG over reference samples drawn from the data distribution, replacing a single baseline with a distribution of baselines~\cite{erion2021expgrad}.
Other methods introduce stochasticity through sampled paths, baseline mixtures, diffusion-based trajectories, or iterative attribution procedures~\cite{lundstrom2022rigorous,barkan2023stochastic,barkan2023iia,lei2024diffusionIG,elisha2024probabilistic}.
These approaches introduce distributions into the attribution procedure, but the model is still evaluated pointwise along sampled input-space trajectories, and the final attribution is obtained by averaging over such trajectories.
Reveal-IG differs by making the object of integration itself an expected model response under a parameterized probe distribution.
The path therefore lives in a lifted space of distributions rather than only in the original input space.

\paragraph{Shapley values and information revelation.}
Shapley-value methods average each feature's marginal contribution over coalitions of observed features~\citep{lundberg2017shap}.
From a path perspective, this corresponds to averaging over discrete feature-reveal paths, where features move one at a time from unobserved to observed.
Reveal-IG keeps this information-revelation interpretation, but replaces discrete feature inclusion with a continuous path of probe distributions.
Related work on additive feature importance clarifies the connection between Shapley values, feature removal, and marginal contributions~\citep{covert2020sage}.

\paragraph{Information-restriction attribution.}
Information bottlenecks for attribution explain a fixed model by injecting noise into intermediate features and identifying the information that must be preserved to maintain the prediction~\citep{Schulz2020IBA,jiang2020IBAEMNLP,Hong2025CoIBA}.
Related information-constrained approaches jointly optimize feature information and prediction, exposing which variables support task-relevant structure in the data~\citep{dib_iclr,dib_pnas}.
Reveal-IG uses a related principle in a complete path-attribution setting: the model is fixed, and attributions integrate changes in expected model response as feature information is revealed continuously from a low-information reference to a high-information endpoint.

\section{Discussion}

Completeness is both the strength and the burden of path-based attribution.
A complete path method must explain the total change between a reference state and the input, but it does so by aggregating all changes encountered along the chosen trajectory.
The central design problem is therefore not only choosing a neutral baseline, but choosing intermediate probes that are relevant to the explanation.
Reveal-IG addresses this by moving path design from input space to distribution space: attribution proceeds by controlled information revelation around the target input.

This distributional view reframes path attribution as a smoothing problem.
Blur-IG smooths the input directly, moving from coarse to fine spatial structure; Reveal-IG instead smooths the model response by evaluating it under probe distributions whose uncertainty decreases along the path.
The attribution therefore depends not only on where the path goes, but on the scale at which the model is queried.
This makes the path a choice of explanatory resolution: early probes ask how the model responds to coarse or uncertain information, while later probes assign credit as the input becomes more precisely specified.

The two-dimensional diagnostics suggest a broader geometric view of attribution.
After factoring out the completeness constraint, attribution fields contain degrees of freedom that reveal method-specific artifacts, such as IG path shadows or SHAP's axis-aligned reveal structure.
These controlled diagnostics complement high-dimensional benchmarks, where such artifacts are harder to isolate.
Across images and tabular data, the same theme recurs: changing the space in which the path is drawn changes which parts of the model's input-output behavior are incorporated into the explanation.
This opens the door to probe families and information-revelation paths matched to the structure of different domains.

\paragraph{Limitations.}
Reveal-IG shifts the main design choice from a baseline input to a family of probe distributions and a path through them.
Different probe families encode different notions of locality, uncertainty, and feature relevance.
Our experiments use factorized Gaussian image probes and factorized empirical tabular probes, which make estimation tractable but ignore correlations among pixels or features.
Reveal-IG is also more expensive than single-gradient methods because it estimates gradients of an expected response over multiple path steps and Monte Carlo samples.
For ResNet-50, IG with 50 integration steps takes $\sim 165$ milliseconds, whereas Reveal-IG's extra 10 Monte Carlo evaluations per integration step brings the time per attribution to $\sim 1.1$ seconds, an increase of about $7 \times$.
In tabular settings, the penalty is far greater: on Bikeshare, it's closer to $1,000 \times$ (IG, KernelSHAP, and LIME are 1-3 milliseconds per attribution, and Reveal-IG takes about 1.5 seconds).
Finally, completeness holds for the lifted function being integrated: finite-variance endpoints decompose the change in expected response under the endpoint distribution, rather than exactly decomposing the deterministic model output.
\medskip
{
\small
\bibliographystyle{unsrtnat}
\bibliography{references}
}


\appendix
 
\section{Implementation Details}
\label{app:impl}
Code for the synthetic, image, and tabular attribution experiments is located at  \href{https://github.com/murphyka/Reveal-IG}{https://github.com/murphyka/Reveal-IG}. 
Hyperparameters are listed in Tables~\ref{tab:hparams_image}\&\ref{tab:hparams_tabular}.

\subsection{Image classification.}
\paragraph{Baselines.}
All baselines use publicly available implementations with their default configurations except where noted.
Vanilla Grad, SmoothGrad, IG, and Expected Gradients are implemented using Captum~\citep{kokhlikyan2020captum}.
IDG uses the authors' public implementation,\footnote{\url{https://github.com/chasewalker26/Integrated-Decision-Gradients/tree/main/IDG}}
and Blur-IG and Guided-IG use the PAIR saliency implementation.\footnote{\url{https://github.com/PAIR-code/saliency/blob/master/saliency/core/blur_ig.py}}\footnote{\url{https://github.com/PAIR-code/saliency/blob/master/saliency/core/guided_ig.py}}
The one exception to default settings is Blur-IG: we use maximum blur scale $\sigma=10$ rather than the PAIR saliency default $\sigma=50$, which performed worse under all reported metrics in our setting.

\paragraph{ImageNet-S Filtering for RMA.}
We use ground-truth segmentation masks from ImageNet-S~\cite{gao2022imagenets}. 
Evaluation is restricted to images (1) for which the model assigns greater than $p$=0.05 to the ground truth class, and (2) have a foreground mask covering more than 2\% of the image area.

\paragraph{Adaptive $\sigma$ selection.}
For Reveal-IG (adaptive), the endpoint noise scale $\sigma_{\text{final}}$ is determined per image by binary search over $[1/256, 1.0]$ to find the largest $\sigma$ satisfying $\mathbb{E}_{\varepsilon}[f(x + \sigma\varepsilon)] \geq 0.95 \cdot f(x)$, where $\varepsilon \sim \mathcal{N}(0, I)$ and the expectation is approximated with 10 samples. This ensures the path endpoint preserves the model's confidence in the target class while allowing image-specific noise tolerance.

\subsection{Tabular regression.}
\paragraph{Tabular model training details.}
For all tabular regression experiments, we use the same MLP architecture:
\[
\mathrm{Linear}(D,64) \rightarrow \mathrm{ReLU} \rightarrow
\mathrm{Linear}(64,64) \rightarrow \mathrm{ReLU} \rightarrow
\mathrm{Linear}(64,1).
\]
Models are trained with Adam using learning rate $10^{-3}$ and batch size 512.
We use a 90/10 train/test split with random shuffling.
All features are treated as continuous variables and are standardized using the training-set mean and standard deviation; regression targets are standardized using training-set target statistics.
The Bike Sharing Demand and California Housing models are trained for 500 epochs; the Wine Quality model is trained for 800 epochs.
Training uses mean-squared error loss for all regression datasets.

\paragraph{Attribution.}
For each evaluation point $x^\star$, we draw a fixed reference pool of 2047 other dataset points, which is used throughout the Reveal-IG attribution calculation. 
For each feature, the initial entropy is computed from the empirical counts of unique feature values in this pool. 
The path is parameterized by feature-wise temperatures $\tau_i$, which define categorical distributions over the pool values through their similarity to $x^\star_i$.
At each path step, each $\tau_i$ is chosen by binary search so that the feature's marginal entropy linearly interpolates between its initial entropy and its endpoint entropy. 
Holding these temperatures fixed within a path step, each Monte Carlo sample constructs a context point by independently sampling one value per feature from these categorical distributions. 
The model is then queried on the sampled context point, and the contribution to the path gradient is obtained using the closed-form derivative of the categorical probabilities with respect to $\tau_i$; no derivative of the model with respect to its input features is required.

Hyperparameters are listed in Tab.~\ref{tab:hparams_tabular}.
All baselines use publicly available implementations with their default configurations except where noted.

\begin{table}[h]
\centering
\caption{Hyperparameters for image attribution methods.}
\label{tab:hparams_image}
\begin{tabular}{lp{8cm}}
\toprule
\textbf{Method} & \textbf{Hyperparameters} \\
\midrule
Vanilla Gradient & Single forward-backward pass \\
SmoothGrad & 50 noise samples, $\sigma_{sg} = 0.15 \times$ input range \\
IG & 50 Gauss-Legendre steps, zero baseline \\
Expected Gradients & 50 samples drawn uniformly from a pool of 100 randomly selected ImageNet training images \\
Guided IG & 200 steps, feature fraction 0.25, max-distance 0.02 (public defaults) \\
IDG & 50 steps, slope-weighted alpha redistribution, zero baseline \\
Blur-IG & 50 steps, max blur $\sigma{=}10.0$ \\
Reveal-IG & 50 integration steps, 10 MC samples per step, $\sigma_\text{final}{=}0.25$ \\
\bottomrule
\end{tabular}
\end{table}

\begin{table}[h]
\centering
\caption{Hyperparameters for tabular attribution methods.}
\label{tab:hparams_tabular}
\begin{tabular}{lp{8cm}}
\toprule
\textbf{Method} & \textbf{Hyperparameters} \\
\midrule
Vanilla Gradient & Single forward-backward pass \\
SmoothGrad & 50 noise samples, $\sigma_{sg} = 0.15 \times$ per-feature training std \\
IG & 64 Riemann midpoint steps, baseline is training set mean \\
Expected Gradients & 64 random training points via Captum \texttt{GradientShap} with \texttt{stdevs} $=0.0$ \\
KernelSHAP & \texttt{n\_samples} $=512$, \texttt{n\_background} $=50$ \\
LIME & \texttt{n\_samples} $=1000$ \\
Feat. Ablation & Captum \texttt{FeatureAblation}, baseline is sample from marginal \\
Reveal-IG & Feature fractional entropy range: $0.99\rightarrow 0.05$, 40 integration steps,  2048 pool size, 40 MC samples per step \\
\bottomrule
\end{tabular}
\end{table}

\section{Ablation experiments}
\label{appx:ablations}

\begin{figure}[htbp]
    \centering
    \includegraphics[width=\linewidth]{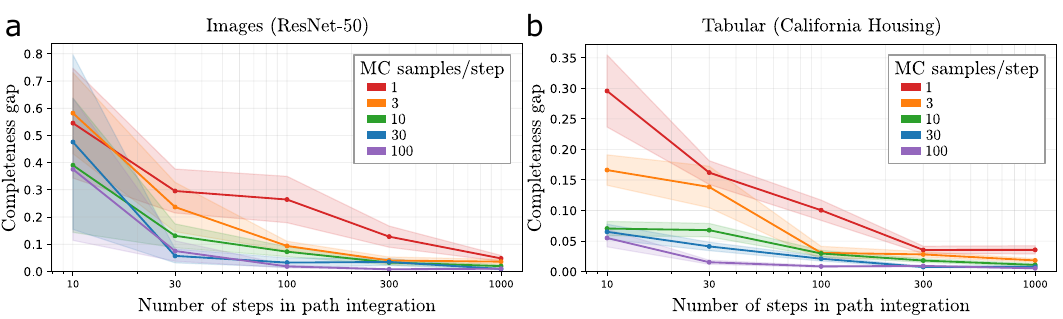}
    \caption{\textbf{Reveal-IG completeness convergence.}
    The completeness gap, measured as $|\sum a_i - (\mathbb{E}_{q_\text{end}}[f(x)]-\mathbb{E}_{q_\text{start}}[f(x)])|$, with $a_i$ the attribution for component $i$, for 
    \textbf{(a)} ResNet-50 on eight ImageNet samples and
    \textbf{(b)} an MLP trained on the California Housing dataset, evaluated for 20 samples.
    Shaded regions display standard error.
    }
    \label{fig:completeness}
\end{figure}

Figure~\ref{fig:completeness} presents ablation studies on the computational requirements to close the completeness gap for Reveal-IG.
Note that the completeness gap, defined as $|\sum a_i - (\mathbb{E}_{q_\text{end}}[f(x)]-\mathbb{E}_{q_\text{start}}[f(x)])|$, compares the sum of attributions across features, $a_i$, to the Reveal-IG path's start and end points.
These will differ in general from the start and end points for other methods: for images, the end point is the expected model response taken over a normal distribution centered at the image and not the image itself as would be the case for IG.
For tabular data, the difference is more apparent in the starting point: while SHAP uses the expectation of the model response taken over the product of feature marginal distributions as the baseline for all datapoints, Reveal-IG uses a broad distribution whose entropy per feature is $0.99\times$ the feature's marginal entropy, resulting in a slightly different baseline value per datapoint.

\subsection{Sigma variation}
\begin{figure}
    \centering
    \includegraphics[width=\linewidth]{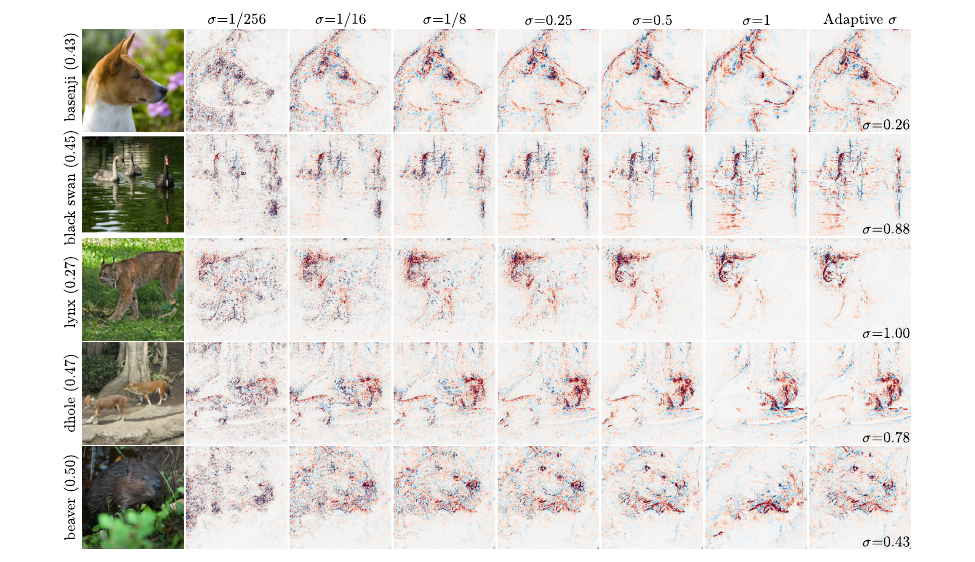}
    \caption{Comparison of Reveal-IG with different $\sigma$ endpoints, including the adaptive (per-image) endpoint, whose values are displayed inside the corresponding attribution maps in the right column.}
    \label{fig:ablation_sigma}
\end{figure}

We compare different $\sigma$ endpoints for Reveal-IG in Tab.~\ref{tab:sigma_ablation} and Fig.~\ref{fig:ablation_sigma}.
The motivation behind the adaptive endpoint is to identify the noise scale at which the model is sensitive to the particular class; $\sigma$ is calculated via binary search over $[1/256, 1.0]$ to satisfy $\mathbb{E}[f(x + \sigma\varepsilon)] \geq 0.95 \cdot f(x)$.
We find that the performance across metrics is best with a fixed endpoint $\sigma=0.25$.
Qualitatively, the attribution maps lose focus when the $\sigma$ endpoint is smaller than around $\sigma$=1/8. 

\begin{table}[t]
\centering
\caption{Reveal-IG $\sigma$ endpoint ablation (ResNet-50, 1k images for Ins/Del/Sens-$n$, ImageNet-S for RMA ($n=879$)).}
\label{tab:sigma_ablation}
\small
\begin{tabular}{lccccc}
\toprule
\textbf{Endpoint $\sigma$} & \textbf{Ins-Del$|$abs $\uparrow$} & \textbf{Ins-Del$|$relu $\uparrow$} & \textbf{RMA $\uparrow$} & \textbf{Sens.-$n$ PCC $\uparrow$} \\
\midrule
$\sigma$=1/256 & 0.1040\ci{0.0066} & 0.1236\ci{0.0060} & 0.5365\ci{0.0172} & 0.0350\ci{0.0042} \\
$\sigma$=0.0625 & 0.1696\ci{0.0077} & 0.1944\ci{0.0082} & 0.5756\ci{0.0159} & 0.0701\ci{0.0043} \\
$\sigma$=0.125 & 0.1854\ci{0.0080} & 0.2171\ci{0.0086} & 0.5850\ci{0.0156} & \second{0.0749}{0.0044} \\
$\sigma$=0.25 & \bgl{0.1960}\ci{0.0081} & \bgl{0.2343}\ci{0.0091} & 0.5956\ci{0.0153} & \bgl{0.0802}\ci{0.0043} \\
$\sigma$=0.5 & \second{0.1956}{0.0082} & \second{0.2270}{0.0089} & \second{0.6060}{0.0153} & 0.0720\ci{0.0042} \\
$\sigma$=1 & 0.1697\ci{0.0082} & 0.1773\ci{0.0083} & 0.5970\ci{0.0164} & 0.0536\ci{0.0041} \\
Adaptive & 0.1912\ci{0.0082} & 0.2193\ci{0.0086} & \bgl{0.6094}\ci{0.0154} & 0.0668\ci{0.0043} \\
\bottomrule
\end{tabular}
\end{table}

\section{Extended synthetic attribution results}
\label{appx:extended_synth}
\begin{figure}
    \centering
    \includegraphics[width=1\linewidth]{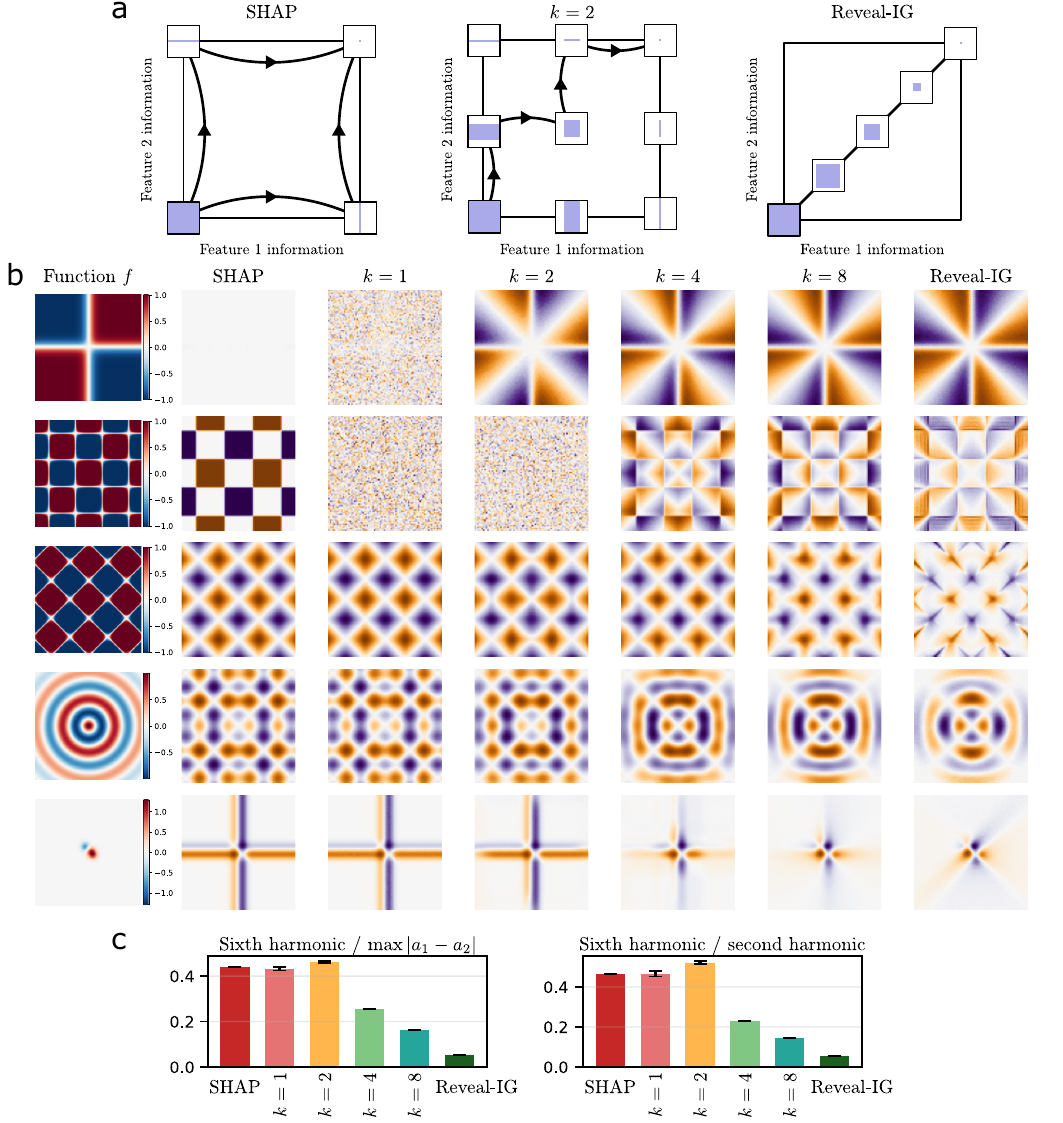}
    \caption{
    \textbf{(a)}
    The manner of feature information revelation compared schematically for SHAP, the subdivided variant discussed in Sec.~\ref{sec:synthetic}, and Reveal-IG.
    For SHAP, the feature contributions are averaged over all paths (here, two possibilities).
    For $k=2$, one path is shown out of the six possibilities.
    The insets' shaded regions visualize the distribution over feature values used for the expectation in that step's calculation, and corresponds to a linear interpolation in the distribution's extent.
    Reveal-IG reveals the feature information along one continuous path.
    \textbf{(b)} For the functions in Fig.~\ref{fig:synthetic_comparison}, we evaluate SHAP, Reveal-IG, and the SHAP analogue.
    Note the attribution heatmaps' color range is $[-b, b]$ with $b=\max(0.10, |a_1-a_2|)$.
    \textbf{(c)} Repeating the azimuthal sweep at radius one as done in Sec.~\ref{sec:synthetic}, the amplitude of the second and sixth harmonics are compared; as $k$ increases, the sixth harmonic decays toward that of Reveal-IG's attribution.
    }
\label{fig:shap_subdivide}
\end{figure}

In Fig.~\ref{fig:shap_subdivide}, we evaluate SHAP, Reveal-IG, and the SHAP analogue where information is revealed in $k$ steps.
Specifically, the functions are defined over the unit square and the data distribution is taken to be uniform over this area.
Whereas SHAP would reveal each feature value entirely in its reveal operation (Fig.~\ref{fig:shap_subdivide}a), the SHAP analogue with $k$ reveal steps gradually shrinks the extent of the uncertainty for a feature during each step.
Here we use a linear interpolation from the full extent to zero over the steps to match Reveal-IG.
We note that once subdivisions are included, choice over revelation path, and multiplicity of valid attributions with it, emerges that is desirably absent from SHAP.

The attribution fields in Fig.~\ref{fig:shap_subdivide}b roughly interpolate between those of SHAP and Reveal-IG, controlled by the number of subdivisions $k$.  
Whereas the simple \texttt{xor} function acquires structure as soon as more than one subdivision is used, the other functions' attributions evolve with $k$.
The angular structure in the radial rings diminishes with $k$ (Fig.~\ref{fig:shap_subdivide}c), as do the line-shadows for the hill-valley near the origin in the final row. 

\section{Qualitative Results}
\label{appx:extra_vis}

\begin{figure}
    \centering
    \includegraphics[width=1\linewidth]{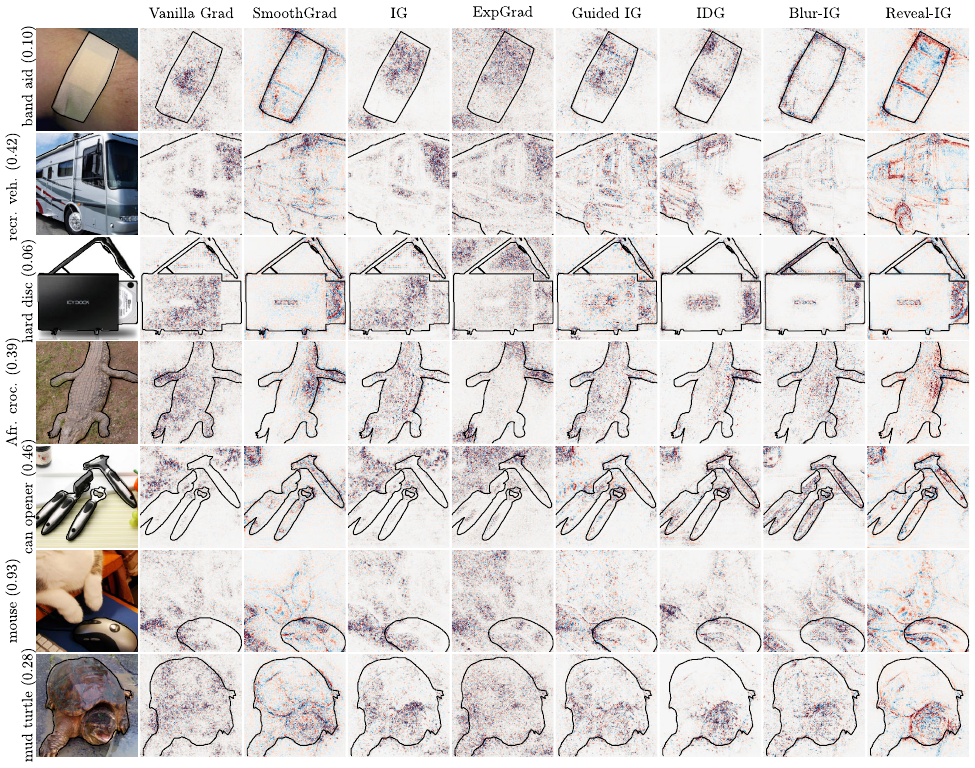}
    \caption{For a random selection of images from the localization set with accompanying ImageNet-S masks and assigned probability to the ground truth class greater than 0.05, we show the outlines of masks in black and the signed attribution maps (red is positive attribution, blue is negative).}
\label{fig:comparison_localization}
\end{figure}

\begin{figure}
    \centering
    \includegraphics[width=\linewidth]{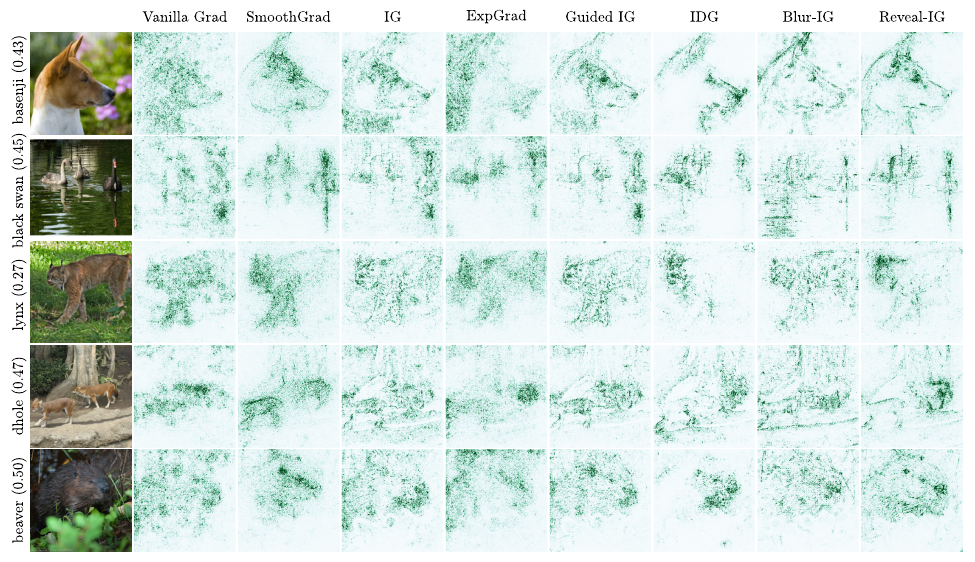}
    \caption{For the same random selection of images from the ImageNet validation set shown in Fig.~\ref{fig:comparison_resnet}, (ViT-B/16 top-1 class and assigned probability on left), we show absolute value saliency maps for the methods compared in Tab.~\ref{tab:image}.}
    \label{fig:comparison_resnet_abs}
\end{figure}

\begin{figure}
    \centering
    \includegraphics[width=\linewidth]{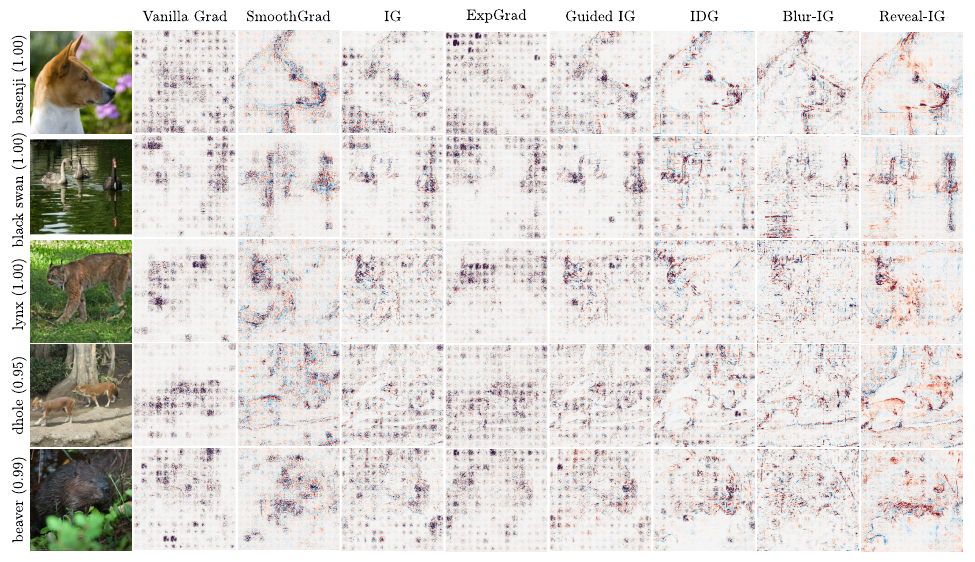}
    \caption{For the same random selection of images from the ImageNet validation set shown in Fig.~\ref{fig:comparison_resnet}, (ViT-B/16 top-1 class and assigned probability on left), we show signed saliency maps for the methods compared in Tab.~\ref{tab:image}. Red is positive attribution and blue is negative.}
    \label{fig:comparison_vit_signed}
\end{figure}

\begin{figure}
    \centering
    \includegraphics[width=\linewidth]{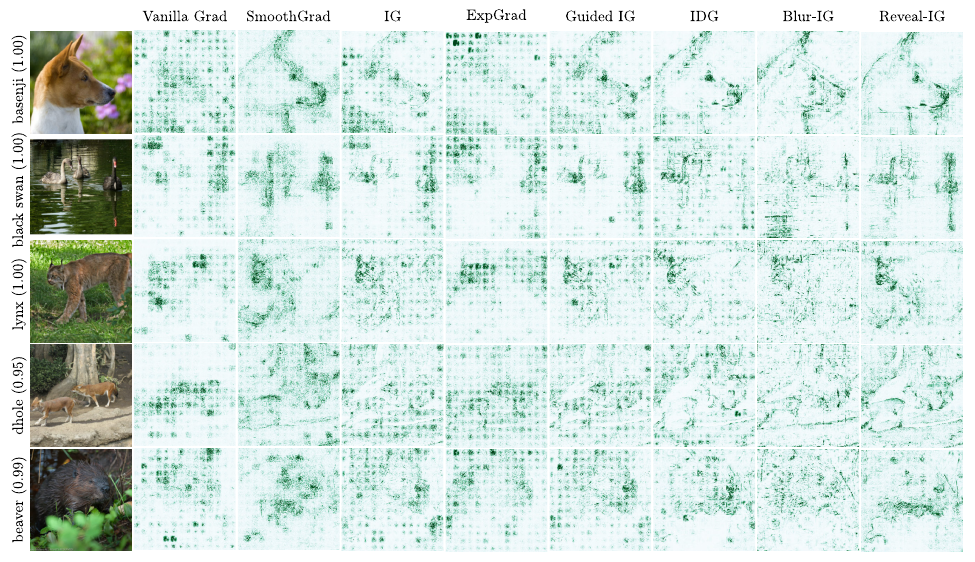}
    \caption{For the same random selection of images from the ImageNet validation set shown in Fig.~\ref{fig:comparison_resnet}, (ViT-B/16 top-1 class and assigned probability on left), we show absolute value saliency maps for the methods compared in Tab.~\ref{tab:image}.}
    \label{fig:comparison_vit_abs}
\end{figure}

\paragraph{Object Image attribution}

The attribution maps Fig.~\ref{fig:comparison_localization} show how each method produces attribution for the image and how much of it concentrates on the object area, denoted by the white boundary. Reveal-IG (adaptive and $\sigma{=}0.25$) keeps most of its attribution mass within the object boundary across diverse classes (Yorkshire terrier, artichoke, whiskey jug, Siamese cat, sea slug), while ExpGrad, Blur-IG, and SmoothGrad spread attribution into the surrounding background regions.

\section{Declarations}

\paragraph{Broader impacts.}
This work develops a feature-attribution method based on distributional paths for information revelation. 
Potential positive impacts include improved tools for interpreting model predictions, diagnosing model behavior, and comparing how different attribution methods assign credit across features. 
Such tools may support model debugging, scientific analysis, and more transparent use of machine-learning systems.

The potential negative impacts arise primarily from misuse or overinterpretation of explanations. 
Attribution methods can produce visually or numerically plausible explanations even when the underlying model is unreliable, biased, or sensitive to the evaluation protocol. 
They may therefore be misused to justify model decisions in high-stakes settings such as healthcare, hiring, lending, or surveillance without sufficient validation. 
Reveal-IG also involves choices of baseline distribution, path, sampling procedure, and perturbation protocol, and these choices can affect the resulting explanation. 
We mitigate these risks by presenting the method as an analysis tool rather than a certification of model reliability, comparing against established attribution baselines, reporting quantitative evaluation results, and explicitly discussing the role of path and baseline choices.

\paragraph{Compute resources.}
Attribution experiments were run on a single workstation with an NVIDIA RTX 4500 Ada GPU. 
The proposed attribution procedure does not require training additional large models. 
For image experiments, attributions were computed using fixed pretrained vision models, with compute dominated by repeated forward and backward passes for each input. 
For tabular experiments, we trained only lightweight predictive models on the evaluated tabular datasets; these models were inexpensive relative to the attribution sweeps. 

\paragraph{LLM usage.}
We used LLM-based programming assistance to help implement and edit experimental code. 
The LLM was not used as a component of the proposed method, was not used to generate experimental results autonomously, and did not determine the experimental design, analysis, or conclusions. 
LLMs were also used for writing support such as copyediting, wording suggestions, LaTeX formatting, and brainstorming.
All code, figures, numerical results, references, and scientific claims were reviewed and verified by the authors. 

\end{document}